\documentclass[10pt,twocolumn,letterpaper]{article}

\usepackage{cvpr}              
\usepackage[accsupp]{axessibility}
\usepackage{graphicx}
\usepackage{amsmath}
\usepackage{amssymb}
\usepackage{booktabs}

\usepackage{booktabs} 
\usepackage{adjustbox}
\usepackage[pagebackref,breaklinks,colorlinks]{hyperref}

\usepackage[capitalize]{cleveref}
\crefname{section}{Sec.}{Secs.}
\Crefname{section}{Section}{Sections}
\Crefname{table}{Table}{Tables}
\crefname{table}{Tab.}{Tabs.}

\def\cvprPaperID{*****} 
\def\confName{CVPR}
\def\confYear{2024}

\title{Emotic Masked Autoencoder on Dual-views with Attention Fusion for Facial Expression Recognition}

\author{Xuan-Bach Nguyen\textsuperscript{1*}, Hoang-Thien Nguyen\textsuperscript{2*},  Thanh-Huy Nguyen\textsuperscript{3}, Nhu-Tai Do\textsuperscript{4}, Quang Vinh Dinh\textsuperscript{5}\\
\textsuperscript{1}Ho Chi Minh City University of Technology, Vietnam\\ 
\textsuperscript{2}Posts and Telecommunications Institute of Technology, Ho Chi Minh City, Vietnam\\
\textsuperscript{3}Ho Chi Minh City University of Education, Vietnam\\
\textsuperscript{4}University of Economics Ho Chi Minh City-UEH Vietnam\\
\textsuperscript{5}Vietnamese-German University, Vietnam\\
{\tt\small bach.nguyenspring@hcmut.edu.vn,
n21dccn080@student.ptithcm.edu.vn}\\
{\tt\small 4401101074@student.hcmue.edu.vn, taidn@ueh.edu.vn, vinh.dq2@vgu.edu.vn}
}
\begin{document}
\maketitle

\begin{abstract}
   \textit{Facial Expression Recognition (FER) is a critical task within computer vision with diverse applications across various domains. Addressing the challenge of limited FER datasets, which hampers the generalization capability of expression recognition models, is imperative for enhancing performance. Our paper presents an innovative approach integrating the MAE-Face self-supervised learning (SSL) method and multi-view Fusion Attention mechanism for expression classification, particularly showcased in the 6th Affective Behavior Analysis in-the-wild (ABAW) competition. By utilizing low-level feature information from the ipsilateral view (auxiliary view) before learning the high-level feature that emphasizes the shift in the human facial expression, our work seeks to provide a straightforward yet innovative way to improve the examined view (main view). We also suggest easy-to-implement and no-training frameworks aimed at highlighting key facial features to determine if such features can serve as guides for the model, focusing on pivotal local elements. The efficacy of this method is validated by improvements in model performance on the Aff-wild2 dataset, as observed in both training and validation contexts.}
\end{abstract}

\section{Introduction}

Facial expressions are key in conveying emotions and are increasingly important in areas like human-computer interaction \cite{liu2017facial}, healthcare \cite{bisogni2022impact}, and driving safety \cite{wilhelm2019towards}. Real-world static image datasets like RAF-DB \cite{li2017reliable}, AffectNet \cite{mollahosseini2017affectnet}, and FERPlus \cite{barsoum2016training} have seen advancements in Facial Expression Recognition (FER) due to deep learning. However, recognizing dynamic facial expressions (DFER) from videos remains challenging due to data scarcity, limited dataset diversity, and the complexity of interpreting expressions over time.



\begin{figure}[t]
  \centering

   \includegraphics[width=\linewidth]{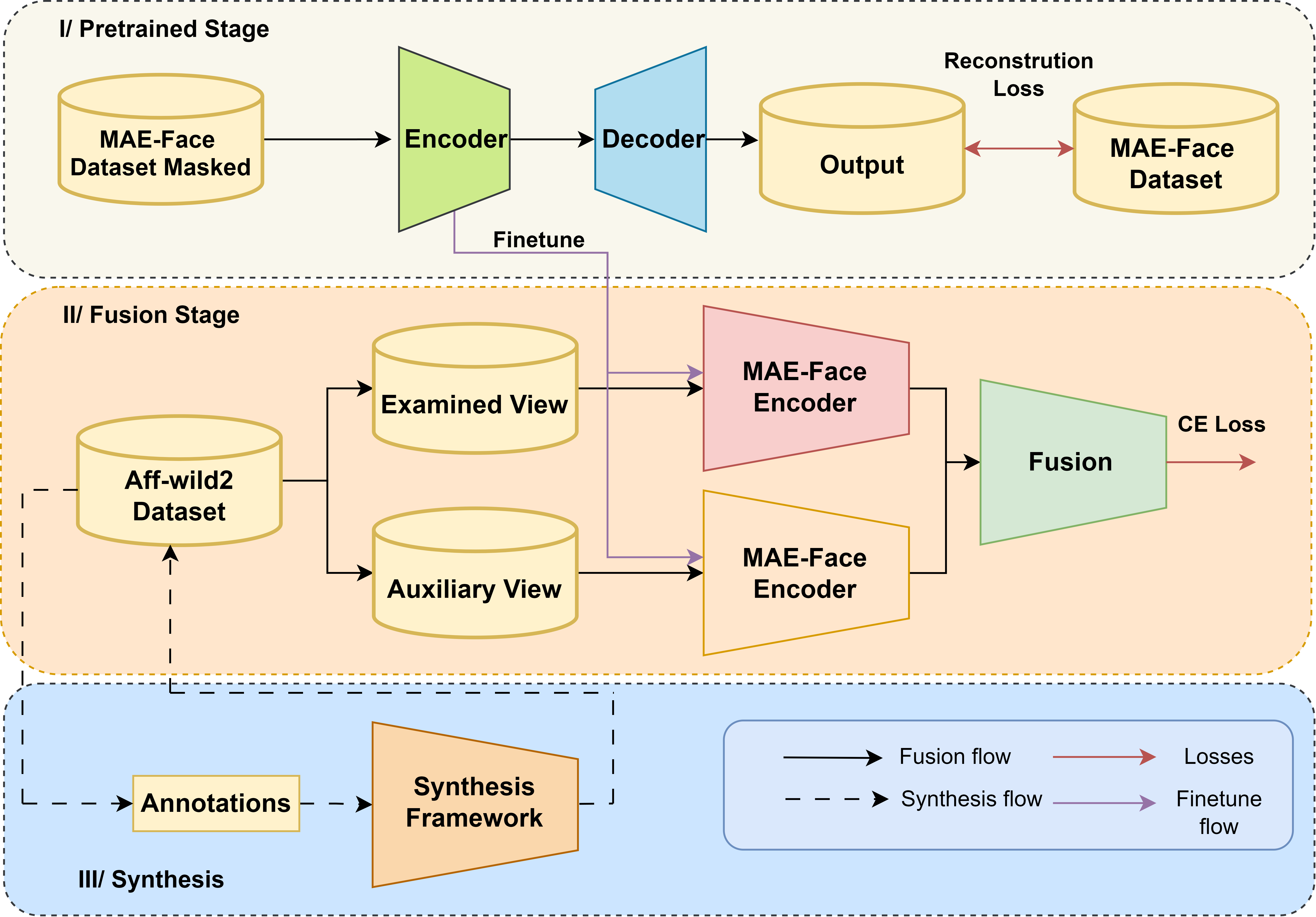}

   \caption{Our proposed pipeline for two-stage pre-training and fine-tuning with fusion, a synthesizing framework to take the informative facial feature with uni-task expression annotations.}
   \label{fig:AU}
\end{figure}
Prior works nowadays usually enhance one path forward by leveraging the ample and varied SFER data as a foundational knowledge base to enhance DFER performance, considering the considerable overlap in information between SFER and DFER data. Additionally, effectively modeling the temporal dimension to capture the nuances of facial expressions over time remains a crucial challenge in DFER. While various methods utilizing 3D \cite{fan2016video} and 2D Convolutional Neural Networks, combined with Recurrent Neural Networks or Transformer \cite{zhao2021former} architectures, have shown promise, these techniques often fall short in explicitly modeling the dynamic nature of facial expressions in videos, thus not fully capitalizing on the temporal information available.

Researchers recently have leveraged Masked Autoencoders (MAE)\cite{li2022affective} to pre-train models on extensive, unlabeled facial datasets, enabling the models to learn the underlying probability distribution of human faces and to reconstruct them effectively. The pre-training phase of the encoder block is thus expected to preserve critical human features that distinguish individual faces, thereby enhancing the model's versatility for a variety of downstream tasks. The work of \cite{ma2023unified} showed that it can improve performance on visual representation by incorporating a two-pass pre-training process and a two-pass fine-tuning process.

Recent studies on facial key points suggest that a focused analysis of facial regions or combining specifically the action units (AUs) \cite{liu2013aware, chen2021understanding,liang2020fine,li2023compound } or feature point tracking algorithm \cite{canedo2019facial}, can enhance the robustness of FER models. This combination is designed to precisely capture facial expressions, thereby directing the model’s attention towards emotion-relevant facial regions for more nuanced analysis. With realizing the potential of action unit facial part by extracting and analyzing cropped AU parts of the face without relying on AU-specific annotations, we shift the focus to recognizing expressions based on overall emotional states rather than discrete facial muscle movements. This approach simplifies the labeling process and potentially reduces the subjectivity in annotations.

Many previous works have been studied about the effectiveness of exploiting multi-view networks to boost the performance of classification tasks \cite{truong2023delving, nguyen2023towards}. Most of these approaches mainly focus on medical applications such as breast cancer diagnosis. Integrating this method with a multi-view analysis capitalizes on the strengths of both static and dynamic expression data. This leads to the creation of a Fusion Multi-View Network that is not only more resilient in interpreting expressions in varied contexts but also reduces the dependency on extensive and precise AU annotations, focusing on the holistic portrayal of emotions. Thus, we aim to address the gaps in Dynamic Facial Expression Recognition (DFER) by building a model grounded in the realistic representation of expressions, reinforced by the insights from the rebuilt smaller static dataset and tailored to the intricacies of dynamic facial expressions by the post-processing. Our approach includes setting up two stages of multi-view analysis. The first stage involves fine-tuning with a pre-trained Masked Autoencoder for Faces (MAE-Face) model, followed by researching various mutations of multi-view networks to understand the effectiveness of fusion at multiple positions in the architecture to robust the feature extracted. We proposed four strategies for fusion types: Mean, Concat, UpDown-Mean, and UpDown-Concat. Overall, our main contributions can be summarized from Fig.\ref{fig:AU} as follows:
\begin{itemize}
    \item \textbf{A new robust facial synthesis auxiliary view framework}: This network uniquely combines cropped AU analysis with feature point tracking, focusing on emotion-related areas with informative regions that are important for emotional representation.
    \item \textbf{Two-Stage Multi-View Analysis with Pretrained MAE-Face Model}: Incorporating a two-stage analysis, starting with fine-tuning a pre-trained MAE-Face model, enhances the network's ability to interpret complex facial expressions by guiding the model to focus on the important image's parts.
    \item \textbf{Comprehensive Study on Fusion Strategies}: By experimenting with four distinct fusion strategies: Mean, Concat, UpDown-Mean, and UpDown-Concat. We investigate the impact of integrating multi-view analysis at different stages in the network architecture, contributing to a more robust feature extraction and interpretation of emotional expressions.
\end{itemize}
\section{Related Work}
\subsection{Facial Expression Recognition}
Facial expression recognition (FER) is a cornerstone of pattern recognition research. Historically, leveraging fully supervised data has marked significant advancements in FER, as evidenced by methodologies detailed in several studies \cite{chang2021learning, chen2021understanding,zheng2023poster,xue2021transfer}. Furthermore, multi-view strategies incorporating dual or multiple images, or supplemental augmented views, have proven to bolster performance \cite{romero2022multi,chen2023static}.
\begin{figure*}[t]
  \centering

   \includegraphics[width=0.95\linewidth]{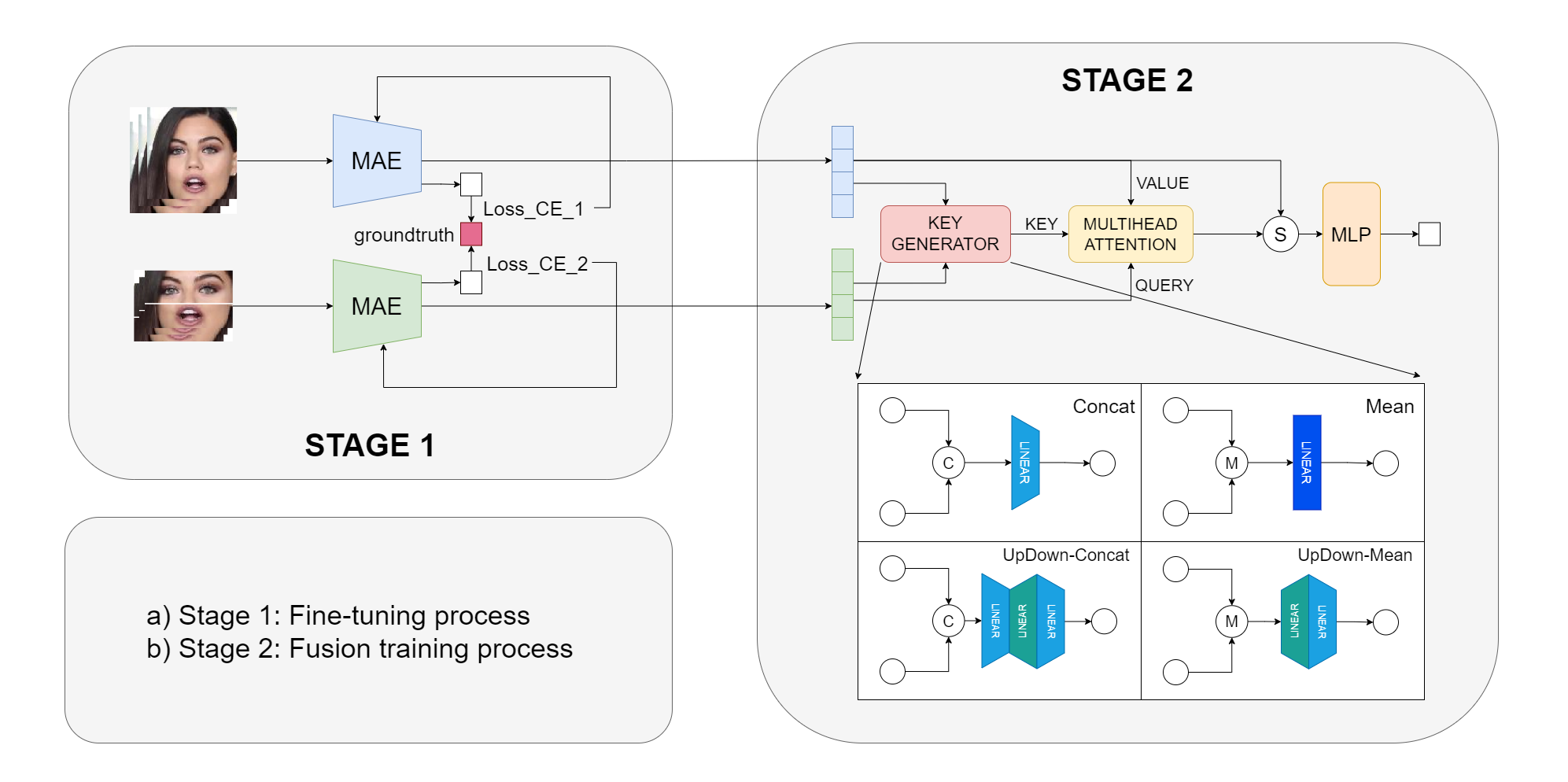}

   \caption{ The architecture of our proposed model consists of two stages. First, the pre-trained model MAE is fine-tuned on two different datasets, the original dataset, and the extracted feature dataset. Secondly, we train the feature spaces of two fine-tuned models on the attention fusion models with four methods, including concat, mean, updown-concat, updown-mean, on the key generator module.}
   \label{fig:model}
\end{figure*}
Recent trends pivot towards fortifying feature connections across divergent feature spaces. One innovative strategy introduces an attention mechanism comprising spatial and channel attention units that collaboratively hone in on pertinent spatial and channel-specific features \cite{wen2023distract}. The POSTER framework extends this concept with an integrated facial landmark detector, a foundational image backbone, cross-fusion transformer encoders, and a pyramid network, addressing inter-class similarity, intra-class variability, and sensitivity to scale in unison \cite{mao2023poster++}. In video applications, the S2D model, a multi-view landmark-aware lightweight adaptation for static images, has set new benchmarks \cite{chen2023static}. The novel MDF-HF architecture encapsulates this evolution, commencing with meticulous data preprocessing. This includes the transformation of video to image sequences, facial detection and cropping, and alignment, thereby minimizing background distractions. Details on adaptive key-frame selection, dynamic feature extraction, and the nuances of hybrid fusion follow \cite{pan2024adaptive}.
\subsection{Learning from unlabeled data} 
Lately, the realm of self-supervised learning has garnered attention for its ability to extract value from unlabeled data, presenting an innovative avenue to mitigate the challenges of annotating FER datasets. These methods\cite{ngo2024dual} diverge from traditional supervised learning, which relies on human-provided labels, by engaging the model in pretext tasks that do not require annotations. Such tasks include image inpainting \cite{pathak2016context}, jigsaw puzzle solving \cite{noroozi2016unsupervised}, and contrastive learning \cite{hadsell2006dimensionality}, among others. For instance, some researchers \cite{li2022affective} utilized Masked Autoencoders (MAE) to pre-train models across extensive facial recognition databases for the ABAW competitions, leading to impressive outcomes. Additionally, a semi-supervised approach introduced by \cite{yu2023exploring} employs a dynamic threshold module (DTM) to effectively harness unlabeled data by adjusting the confidence threshold across various classes and throughout different stages of training.

Nonetheless, it's important to note that these pretext tasks are primarily designed for general image classification purposes, such as identifying object species, and may not be as effective in directly extracting features that are critical for recognizing facial expressions in FER tasks.
\section{Methodology}

\label{sec:intro}

In this section, we will describe our proposed approach in detail in Fig.\ref{fig:model}. At the core of our model lies an architecture composed of several critical components. Initially, we employ a pre-trained Masked Autoencoder (MAE-Face) as a primary feature extractor to capture nuanced facial expressions and affective cues. This base is further refined by fine-tuning the MAE-Face on the Aff-wild2 dataset, ensuring our feature extraction is highly tailored to the intricacies of affective behavior. To process these features effectively, we incorporate self-attention blocks, which allow our model to discern the most relevant affective signals dynamically. The integration of these elements culminates in a Multilayer Perceptron (MLP) that serves as the final prediction mechanism, offering a precise and reliable analysis of emotional expressions.


\subsection{Pre-processing}

\subsubsection{Clean data}
First of all, in order to uphold the integrity and quality of the training dataset, we implement a rigorous data-cleaning process. This involves the systematic removal of corrupt files, non-face images, and those with insufficient resolution, thereby ensuring the robustness and reliability of the dataset for subsequent analyses and model training.
\subsubsection{Synthesis Framework}
\begin{figure*}[ht]
  \centering

   \includegraphics[width=\linewidth]{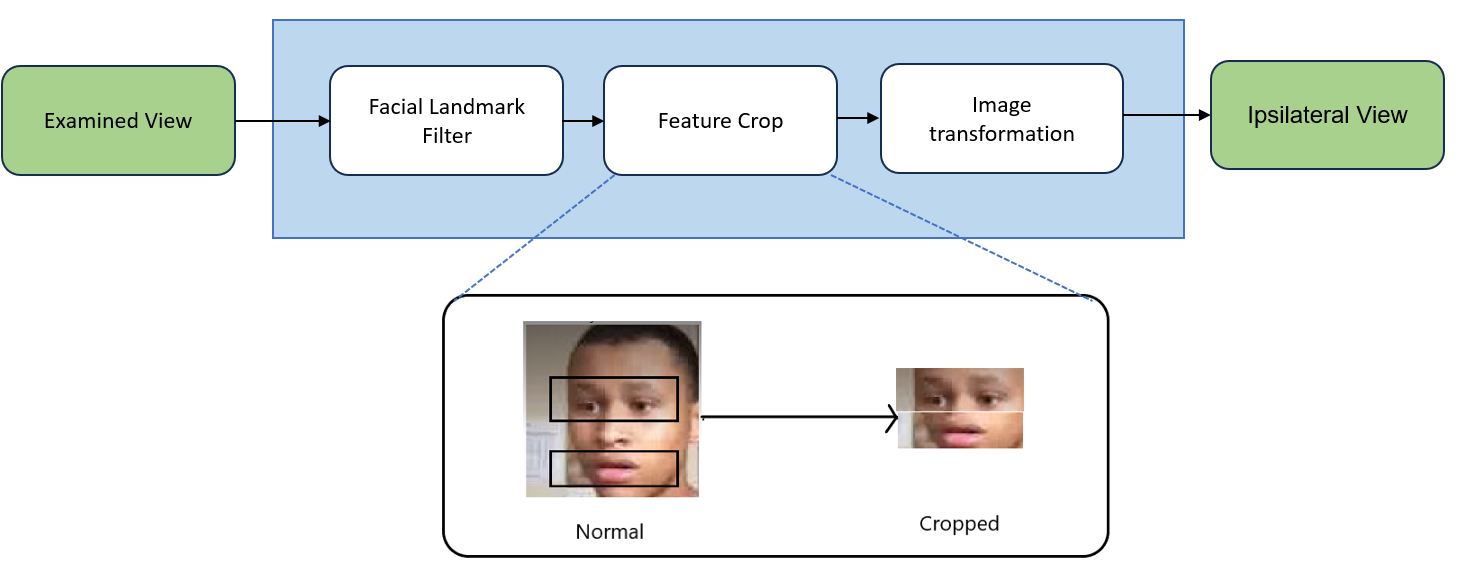}

   \caption{ After the images are cropped and aligned in the Aff-wild2 dataset, the image parts that contain just the mouth and eye are extracted for further processing.}
   \label{fig:model2}
\end{figure*}

The Dlib framework \cite{king2009dlib} is applied to filter out all the facial images that do not have sufficient key points for eyes, eyebrows, nose, and mouth. In the fine-tuning phases, face images are cropped by a percentage of frame sizes to extract features of the eyes and mouth. The eyes are extracted by cutting from 0.35 to 0.55 in height and 0.2 to 0.8 in width, and the mouth is cropped with the same length but at 0.7 to 0.9 in height Fig.\ref{fig:model2}. Finally, image transformation is applied to resize the face image patches.

\subsection{MAE-Face}
The proposed model employs a Vision Transformer (ViT) pre-trained via a self-supervised learning method, leveraging a Masked Autoencoder (MAE) for the process called MAE-Face\cite{ma2022facial}. Specifically, input images are segmented into 16$\times$16 patches, of which 75\% are masked, focusing the training on reconstructing these patches from the remaining visible ones. The model is pre-trained on a comprehensive facial image dataset, including AffectNet \cite{mollahosseini2017affectnet}, CASIA-WebFace\cite{yi2014learning}, IMDB-WIKI\cite{rothe2015dex}, and CelebA\cite{zhang2020celeba}, totaling 2,170,000 images, to enhance its capability in understanding facial features without relying on labels. Post pre-training, the model transitions to fine-tuning for downstream task Expression recognition on the Aff-wild2 dataset. We extracted the feature based on the pre-training and fine-tuned the MAE face for further processing.

\subsection{Fusion Attention block for Emotion Recognition}
In the approach employed, a fusion-based methodology is utilized for enhancing facial emotion recognition (FER) through the integration of two pre-trained models. The methodology makes use of an attention-based network, concentrating on both self-attention and local attention, to effectively combine the strengths of these models.

\subsubsection{Fusion Attention Network}
The proposed fusion attention network incorporates a multi-layer perceptron (MLP) for combining features from two emotion recognition models. By concatenating or adding and then downsampling these features, the network applies self-attention and local attention mechanisms in the Multihead Attention Block to refine the feature representation for emotion classification.

\subsubsection{Self-Attention and Local Attention}
The self-attention mechanism operates by computing the attention scores based on the dot product of features from pre-trained networks, serving as the Key ($K$) and Query ($Q$), followed by a softmax function. The local attention mechanism employs convolution operations to extract more refined features. The attention formula is defined as:

\begin{equation}
\text{Attention}(Q, K, V) = \text{softmax}\left(\frac{QK^T}{\sqrt{d_K}}\right) \otimes V
\end{equation}

where $K$ is the Key, $Q$ is the Query, $V$ is the Value, and $d_K$ is the dimensionality of the Key. This approach not only leverages the distinct advantages of both attention mechanisms but also enhances the model's ability to recognize complex emotional states from facial expressions.

\subsubsection{Skip Connection}
To bolster the model further, a skip connection is introduced to preserve and reinforce prominent features, thereby enhancing the learning process and robustness of the model. The skip connection allows for the direct flow of information by adding the input directly to the output of the attention block. The formula for a skip connection, represented in LaTeX code, is as follows:

\begin{equation}
\text{Output} = \text{Attention}(Q, K, V) + \text{X}
\end{equation}

Where $x$ represents the stronger input feature. This addition of the input to the output of the attention mechanism helps in mitigating the issue of vanishing gradients, allowing for deeper network architectures without the loss of relevant feature information through the layers.
\subsection{Loss function}
The cross-entropy loss function is commonly used in classification tasks and calculates the difference between the predicted probability distribution and the true distribution of the classes.

The standard cross-entropy loss function is given by:

\begin{equation}
\mathcal{L} = -\sum_{i=1}^{N} \sum_{j=1}^{C} y_{ij}\log(\hat{y}_{ij})
\end{equation}
where $N$ is the number of batch samples, $C$ is the number of predicted classes (eight for Aff-Wild2), and $y_{ij}$, $\hat{y}_{ij}$ represent the ground-truth label and predicted scores, respectively.

\subsection{Post-processing}
Given that the Aff-Wild2 dataset is derived from the entirety of video frames and facial expressions typically evolve gradually over time, rapid changes in expressions across adjacent frames are uncommon. Furthermore, following the removal of non-face, corrupted, and low-resolution images, a sliding window approach is adopted for post-processing prediction results to ensure smoother labels. We chose a window size is 50 for our approach.\\

\section{Experiment}
\label{sec:formatting}

In this section, we will provide a detailed description of the used datasets, the experiment setup, and the experimental results.


\subsection{Dataset}
\begin{table*}[ht]
  \centering
 \small
  \begin{tabular}{|l|c|cccccccc|c|}
    \hline
    Features & Accuracy & Neutral & Anger & Disgust & Fear & Happy & Sad & Surprise & Other & Marco F1\\
    \hline
    Not crop (normal) & \textbf{0.512} & \textbf{0.624} & \textbf{0.295} & \textbf{0.188} & 0.015 & \textbf{0.515} & \textbf{0.448} & \textbf{0.287} & 0.502 & \textbf{0.359} \\
    \hline
    Eye & 0.257 & 0.390 & 0.001 & 0.000 & \textbf{0.412} & 0.134 & 0.012 & 0.052 & 0.213 & 0.157 \\
    \hline
    Nose & 0.239 & 0.071 & 0.049 & 0.117 & 0.033 & 0.244 & 0.121 & 0.063 & 0.402 & 0.137 \\
    \hline
    Mouth & 0.324 & 0.388 & 0.072 & 0.011 & 0.406 & 0.250 & 0.246 & 0.110 & 0.433 & 0.239 \\
    \hline
    Eye + Mouth & 0.491 & 0.559 & 0.257 & 0.023 & 0.019 & 0.482 & 0.233 & 0.257 & \textbf{0.559} & 0.304 \\
    \hline
    Eye + Mouth + Nose & 0.443 & 0.551 & 0.192 & 0.015 & 0.233 & 0.512 & 0.289 & 0.165 & 0.495 & 0.302 \\
    \hline
  \end{tabular}
  \caption{Results on fine-tuning process of the validation set off Aff-widl2}
  \label{tab:finetune}
\end{table*}

\begin{table*}[ht]
  \centering
 \small
  \begin{tabular}{|l|c|cccccccc|c|}
    \hline
    Fusion & Accuracy & Neutral & Anger & Disgust & Fear & Happy & Sad & Surprise & Other & Marco F1\\
    \hline
    Mean & 0.527 & 0.615 & 0.349 & 0.231 & 0.012 & 0.527 & 0.454 & 0.264 & 0.553 & 0.376 \\
    \hline
    Concat & \textbf{0.529} & \textbf{0.622} & \textbf{0.364} & 0.241 & 0.018 & \textbf{0.538} & 0.432 & \textbf{0.271} & \textbf{0.554} & \textbf{0.380}\\
    \hline
    UpDown + Mean & 0.518 & 0.613 & 0.282 & 0.265 & 0.018 & 0.524 & 0.453 & 0.252 & 0.539 & 0.368\\
    \hline
    UpDown + Concat & 0.520 & 0.614 & 0.261 & \textbf{0.274} & \textbf{0.033} & 0.520 & \textbf{0.457} & 0.268 & 0.549 & 0.372\\
    \hline
  \end{tabular}

  \caption{Results on fusion process of the validation set of Aff-wild2}
  \label{tab:fusion}
\end{table*}
Aff-wild2 \cite{kollias20246th,kollias2023abaw2,kollias2023multi,kollias2023abaw,kollias2022abaw,kollias2021analysing,kollias2021affect,kollias2021distribution,kollias2020analysing,kollias2019expression,kollias2019deep,kollias2019face,zafeiriou2017aff} is a large-scale video database for ABAW competitions. It annotated 548 videos, around 2.7M frames, into eight pre-defined categories: anger, disgust, fear, happiness, sadness, surprise, neutral, and others. Thanks to the release of this database, we conduct experiments to explore the effectiveness of our method in the ABAW challenge. In our paper, we obtain 8,000 labeled images for each class category from the Aff-Wild2 dataset through uniform sampling.

\subsection{Setup}
All training face images are resized to 224×224 pixels,
our proposed method is implemented with the PyTorch toolbox on NVIDIA Tesla A6000 GPUs. The model is fine-tuned with the Adam optimizer for 100 iters. The learning rate for the fine-tuning process is set at 1e-4. The batch size is set to 512. The average f1 Score across all eight categories on the validation set is reported.
\subsection{Metrics}
The average F1 Score across all eight categories on the validation set is measured as a performance assessment.
\begin{equation}
P = \sum_{i=1}^{8} \frac{F1_i}{8}
\end{equation}
where \( F1 \) denotes f1-score, is calculated by:
\begin{equation}
F1 = 2 \cdot \frac{\text{precision} \cdot \text{recall}}{\text{precision} + \text{recall}}
\end{equation}
\subsection{Results}
In the fine-tuning process, we experience many methods with our approach as shown in Table \ref{tab:finetune}. We have tried training the model with just eyes, nose, or mouth. After that, we try other cases by combining these objects, such as eye and mouth or eye and mouth and nose, to analyze the importance and influence of each feature on the emotions. However, we prioritize the normal feature model, which remains the original image, and the eye-mouth model, which uses the eye and mouth of the face, to feed into the fusion training process. The ratio gets each feature:
\begin{itemize}
    \item Eye: width $\in$ [0.2, 0.8], height $\in$ [0.35, 0.55]
    \item Mouth: width $\in$ [0.2, 0.8], height $\in$ [0.7, 0.9]
    \item Nose: width $\in$ [0.4, 0.6], height $\in$ [0.2, 0.8]
\end{itemize}

Next, we utilize the feature embeddings of two distinct models to train the fusion attention model. We change the key generator in many cases, such as concat methods (includes 1 dense layer), mean method (1 dense layer), UpDown-Mean(2 dense layers), and UpDown-Concat (3 dense layers). Finally, we conclude that the concat method is the main approach of our model. We can see the results in Table \ref{tab:fusion}.

\subsection{Ablation Studies}
An ablation study was conducted to assess the contribution of different facial regions and fusion methods to the performance of a facial expression recognition model. The study's results are delineated in two tables, presenting the outcomes of the fine-tuning process and the fusion strategies, respectively.
\begin{table}[ht]
  \centering
 \small
  \begin{tabular}{|l|c|}
        \hline
   Method & F1\\

        \hline
    w/o. MAE finetune, Fusion, Post-processing  & 0.236 \\
        \hline
    w/o. Fusion,Post-processing  & 0.359 \\
    \hline
    w/o. Post-processing & 0.380 \\
    \hline
    Ours & 0.401 \\
      \hline
  \end{tabular}

  \caption{Results of overall method of the validation set of Aff-wild2}
  \label{tab:final}
\end{table}
From the first table, it's evident that different facial features contribute variably to the accuracy of the model. When the model does not use any cropped features and utilizes the full face (normal), it achieves a baseline accuracy of 0.512. However, when isolating the features, the 'Eyes' alone result in a lower accuracy (0.257), which suggests that while the eyes are important for recognizing some emotions, they are not sufficient on their own however they can highlight where the expression is 'Fear'. The 'Nose' and 'Mouth' features also show limited effectiveness when used independently, with accuracies of 0.239 and 0.324, respectively.

Combining features improves the model’s performance, with 'Eye + Mouth' significantly improving the accuracy to 0.491. The combination of all three features ('Eye + Mouth + Nose') further enhances the performance, yielding an accuracy of 0.443. This implies that a holistic representation of facial features, encompassing multiple regions, is more effective for emotion recognition tasks.

The second table explores the effectiveness of different feature fusion strategies. 'Mean' fusion leads to an accuracy of 0.527, while 'Concat' fusion slightly outperforms it with an accuracy of 0.529, suggesting that concatenation of features might be slightly more beneficial than averaging them. Furthermore, advanced fusion strategies, such as 'UpDown + Mean' and 'UpDown + Concat', yield comparable results with accuracies of 0.518 and 0.520, respectively. These methods appear to be robust, offering a balance between detail retention and abstraction.

As shown in Table \ref{tab:final}, by implementing a sliding window approach with a window size of 50, a technique informed by the study in \cite{savchenko2023emotieffnet}, our models attained an overall F1 macro score of 0.401. This indicates that selecting an optimal smoothing window size, denoted as \textit{k}, can significantly enhance the performance metrics observed on the validation dataset.

The performance of our method in the 6th ABAW challenge as shown in Table \ref{Final-table}, while not reaching the high mark set by the leading team, Netease Fuxi AI Lab, does manage to exceed the baseline. This indicates a promising direction for our approach, suggesting that with further development, there's potential to bridge the gap to the SOTA performance.
\begin{table}[ht]
\centering
\begin{tabular}{lcc}
\hline
\textbf{Teams} & \textbf{Test} \\
\hline
Netease Fuxi AI Lab & \textbf{0.5005} \\
CtyunAI & 0.3625 \\
USTC-IAT-United & 0.3534 \\
HSEmotion & 0.3414 \\
M2-Lab-Purdue & 0.3228 \\
KBS-DGU & 0.3005 \\
SUN-CE & 0.2877 \\
\textbf{Ours} & 0.2797 \\
CAS-MAIS & 0.265 \\
IMLAB & 0.2296 \\
baseline & 0.2250 \\
\hline
\textbf{Ours (Post challenge)} & 0.2866\\
\end{tabular}
\caption{Result on the test set of 6th ABAW challenge}
\label{Final-table}
\end{table}

\section{Conclusion}

In this study, we present an efficient approach for evaluating the performance of the Fusion MAE-Face technique in uni-task expression analysis using the Aff-Wild2 dataset. Our method improves the learning of facial expressions by refining preprocessing steps that focus on facial features while reducing interference from background pixels. The model initially learns low-level features from two symmetric perspectives separately and then combines these features to learn high-level information through a process of feature fusion. Our tests on the Aff-Wild2 dataset show that starting with pre-trained MAE-Face weights and integrating fusion attention mechanisms significantly enhances the model's ability to identify meaningful features. This allows our method to outperform baseline models with less than 10\% of the training data. Our methodology is innovative and provides new insights into focusing on crucial local details and exploring how different views can enhance model performance.

\section*{Acknowledgement}
This research was funded by University of Economics Ho Chi Minh City, Vietnam and AI VIETNAM.

{\small
\bibliographystyle{ieee_fullname}
\bibliography{egbib}
}

\end{document}